%% file: main.tex
\documentclass[runningheads]{llncs}

\usepackage[mobile]{eccv}

\input{preamble}

\usepackage{eccvabbrv}

\usepackage{graphicx}
\usepackage{booktabs}

\usepackage[accsupp]{axessibility}  %

\usepackage[pagebackref,breaklinks,colorlinks,citecolor=eccvblue]{hyperref}

\usepackage{orcidlink}

\begin{document}

\title{\methodname: Multi-Subject Action Binding in Generative Video Games} 

\titlerunning{\methodname}

\author{
Alexander Pondaven\inst{1,2} \thanks{Work done during internship at Snap Inc. Corr: \email{pondaven@robots.ox.ac.uk}} \and
Ziyi Wu\inst{3} \and
Igor Gilitschenski\inst{3} \and
Philip Torr\inst{2} \and
Sergey Tulyakov\inst{1} \and
Fabio Pizzati\inst{4} \and
Aliaksandr Siarohin\inst{1}
}

\authorrunning{A. Pondaven et al.}

\institute{
Snap Research \and
University of Oxford \and
University of Toronto \and
MBZUAI
}

\maketitle

\begin{center}
    \href{https://action-party.github.io/}{action-party.github.io}
\end{center}

\input{sec/0_abstract}

\input{sec/1_intro}

\input{sec/2_related}

\input{sec/3_preliminaries}
\input{sec/4_method}

\input{sec/5_experiments}

\input{sec/6_conclusions}

\input{sec/7_acknowledgements}

\bibliographystyle{splncs04}
\bibliography{main}

\input{sec/X_suppl}

\end{document}

%% file: preamble.tex
\usepackage{algorithm}
\usepackage{mathtools}
\usepackage{mdframed}
\usepackage[export]{adjustbox}
\usepackage{wrapfig}
\usepackage{multirow}
\usepackage{enumitem}
\usepackage{algorithmic}

\newcommand{\methodname}{ActionParty\xspace}

\newcommand{\heading}[1]{\vspace{.25em}\noindent\textbf{#1}}

\newlength\pagetopmargin
\newlength\figcapmargin
\newlength\figmargin
\newlength\tablecapmargin
\newlength\tablemargin

%% file: sec/0_abstract.tex
\begin{abstract}

  Recent advances in video diffusion have enabled the development of ``world models'' capable of simulating interactive environments.
  However, these models are largely restricted to single-agent settings, failing to control multiple agents simultaneously in a scene.
  In this work, we tackle a fundamental issue of action binding in existing video diffusion models, which struggle to associate specific actions with their corresponding subjects.
  For this purpose, we propose \methodname, an action controllable multi-subject world model for generative video games.
  It introduces \emph{subject state} tokens, \textit{i.e.} latent variables that persistently capture the state of each subject in the scene.
  By jointly modeling state tokens and video latents with a spatial biasing mechanism, we disentangle global video frame rendering from individual action-controlled subject updates.
  We evaluate \methodname on the Melting Pot benchmark, demonstrating the first video world model capable of controlling up to seven players simultaneously across 46 diverse environments.
  Our results show significant improvements in action-following accuracy and identity consistency, while enabling robust autoregressive tracking of subjects through complex interactions.

  \keywords{Video generation \and World models \and Action binding} 
  
\end{abstract}

%% file: sec/1_intro.tex
\begin{figure}[!t]
  \centering
  \includegraphics[width=\textwidth]{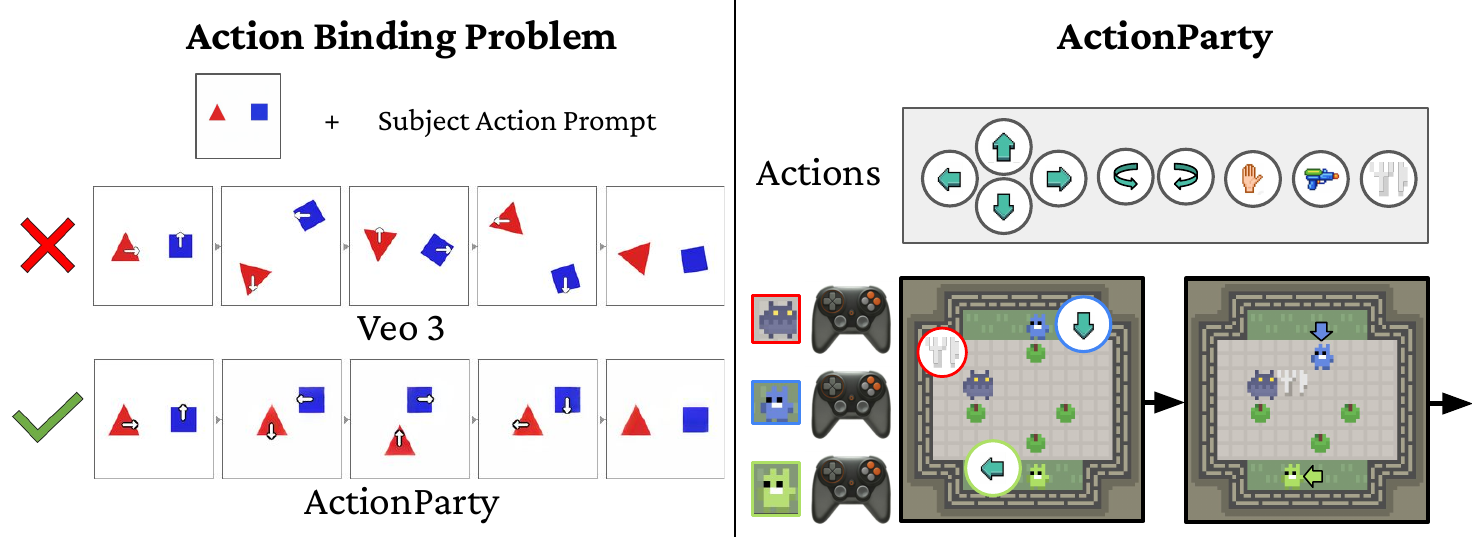}
  \caption{Left: Action binding failure case of text-to-video models with prompt: `The red triangle moves right and the blue square moves up. Then the red triangle moves down and the blue square moves left. Then the red triangle moves up and the blue square moves right. Then the red triangle moves left and the blue square moves down.' Right: \methodname enables action control of multiple subjects in a scene.
  }
  \label{fig:teaser}
  \vspace{\figmargin}
\end{figure}

\section{Introduction}
\label{sec:intro}

The emergence of video diffusion models~\cite{DDPM, VideoDiffusionModels, Wan21, ImagenVideo, SVD, Sora} has shifted the focus of generative AI from static content creation to the development of ``world models'', simulators capable of predicting future observations conditioned on interactive actions~\cite{bruce2024geniegenerativeinteractiveenvironments, RELIC, ha2018worldmodels}.
By learning the underlying dynamics of environments directly from pixels, recent breakthroughs such as Genie~\cite{genie3} and GameNGen~\cite{GameNGen} have demonstrated the potential for generating high-fidelity, interactive video games.
Despite these advances, existing world models are often restricted to single-agent settings with an egocentric output view~\cite{genie3}.
They accept as input a single stream of control signals and can only control a single subject in the scene.
Consequently, these generated ``worlds'' are unable to simulate the multi-agent dynamics that define complex social environments such as in self-driving~\cite{GAIA-1, TrafficSim} and robotics~\cite{VMAS, RoboTwin2} applications.

In this work, we aim to explore multi-subject control in video world models.  %
Since video diffusion models can act as world models, a natural adaptation of existing control mechanisms is to use textual instructions describing the actions of all actors as input to the model. However, this approach leads to a catastrophic failure in action-subject association: the model fails to apply an action to the correct target subject. 
To diagnose this, we analyze a simplified environment consisting of two primitive shapes moving on a solid background in \cref{fig:teaser}.
Surprisingly, even state-of-the-art video generators like Veo 3~\cite{GoogleVeo} struggle to execute a sequence of basic commands such as ``triangle moves down while square moves left''. Failures are even more significant when several actions are described in sequences, as we show in the figure. This failure relates to a fundamental limitation of attribute binding in existing diffusion models. Indeed, prior work shows that when given multiple conditioning signals, diffusion models often ignore some of them or merge multiple signals together~\cite{LLMGroundedDM, GLIGEN, MinT}, preventing them from achieving per-subject control required by multi-agent simulation.

As a stepping stone for solving the action-binding issue, we propose \methodname, a multi-subject world model focusing on multi-player game environments, where each player is a controllable subject in a video generation fashion.Inspired by recent works that provide explicit spatio-temporal grounding as model input~\cite{MinT, GLIGEN, LLMGroundedDM}, we introduce \emph{subject state} tokens, a set of latent variables that serve as a persistent implicit state for each subject in the scene. 
Ultimately, \methodname operates as a generative game engine, jointly modeling subject state tokens and video latents.Subject state tokens are integrated using an attention mask that enforces action-subject correspondence. Moreover, to ensure actions are applied to the correct subjects in the video, we leverage 3D Rotary Position Embeddings (RoPE)~\cite{su2023roformerenhancedtransformerrotary} to bias subject tokens to the current spatial locations of subjects within the video.

We evaluate \methodname on video outputs from the Melting Pot benchmark~\cite{agapiou2023meltingpot20}, a challenging suite of 46 2D multi-agent games.
We learn a unified action space that applies to all games, and our model maintains precise per-subject control over up to seven players.
Thanks to the separate modeling of subject states, we significantly outperform text-only baselines in action-following accuracy and identity consistency.
Beyond pure video generation, we show that the co-generated subject states accurately follow the moving subjects in the video, even through complex interactions and changes in character states.
In summary, we make the following contributions:

\begin{itemize}[noitemsep,topsep=0pt]
    \item We propose \methodname, a novel autoregressive video generator for generative game simulation.
    We jointly model video frames and the state of each subject and introduce attention masking to enforce action-subject correspondence to enable consistent multi-subject action control.
    \item We introduce an action binding mechanism based on RoPE biasing that anchors subject state to specific spatial coordinates within the video, enabling reliable subject localization.
    \item We demonstrate the first video world model capable of controlling up to seven subjects simultaneously across 46 distinct environments, achieving superior performance in action-following and subject consistency.
\end{itemize}

%% file: sec/2_related.tex
\section{Related work}

\heading{Video diffusion models.}
Since the rise of diffusion models~\cite{DiffusionModels, DDPM}, text-to-video and image-to-video generation have achieved tremendous progress by training on Internet-scale data~\cite{VideoDiffusionModels, ImagenVideo, SVD, DynamiCrafter, HunyuanVideo, Step-Video-T2V, Seaweed-7B}.
While earlier approaches often adopt U-Net~\cite{U-Net} as the denoising network~\cite{AlignYourLatents, AnimateDiff, VideoCrafter1, VideoCrafter2}, recent models have switched to the diffusion transformer (DiT)~\cite{peebles2023scalablediffusionmodelstransformers} architecture due to their better scalability in generating high-resolution and complex videos~\cite{Sora, Wan21, SnapVideo, CogVideoX, WALT, MetaMovieGen}.
Conventional video DiTs typically denoise all video frames jointly.
To achieve real-time interactive video generation, later works study distilling a pre-trained bi-directional DiT to a \emph{causal} autoregressive DiT~\cite{CausVid, he2025matrixgame20opensourcerealtime, yu2025gamefactorycreatingnewgames, RELIC}.
This often involves Diffusion Forcing~\cite{chen2024diffusionforcingnexttokenprediction} that trains the model with an independent noise level per frame, followed by Self-Forcing~\cite{SelfForcing} that fine-tunes the model on self-generated rollout with a distribution matching loss~\cite{DMD, DMDv2}.
In this work, we follow similar techniques to train an autoregressive action-conditioned video DiT.

\heading{Video world models.}
World models~\cite{ha2018worldmodels, ha2018recurrent} are generative models that are trained to predict the next state (e.g., visual observation) given an action.
Actions may be defined as any conditioning signal, such as discrete arrow keys in a video game or text descriptions of changes in a scene.
The playable video line of work~\cite{menapace2021playablevideogeneration, menapace2022playableenvironmentsvideomanipulation} was one of the first in this direction.
They try to learn latent actions from raw videos via inverse dynamics.
Another line of work simulates first-person view games like Doom~\cite{valevski2024diffusionmodelsrealtimegame}, CSGO~\cite{DDPMWorldModel}, Minecraft~\cite{oasis2024, WorldMem}, and more~\cite{DDPMWorldModel, zhang2025matrixgameinteractiveworldfoundation, he2025matrixgame20opensourcerealtime, yu2025gamefactorycreatingnewgames} where actions are recorded during gameplay.
The recent Genie~\cite{bruce2024geniegenerativeinteractiveenvironments, genie3} line of research further scales training to real-world videos with both text descriptions and keyboard inputs as actions.

While existing world models excel at single-actor control, there is little work extending them to multiplayer settings, where separate actions must be applied to multiple subjects within a scene.
Multiverse~\cite{enigma2025multiverse} and Solaris~\cite{solaris2026} generate aligned views for each subject in a 3D world.
They model each actor's view individually, making action binding trivial.
However, separate actor modeling scales the number of video tokens linearly with the number of actors, which may explain why they only test on two-player settings in a single game.
MultiGen~\cite{multigen} is concurrent work that also predicts explicit game state (including a minimap and player states) for aligning multiplayer views. Existing work apply one action to individual player video streams.
In contrast, our method generates one video containing all actors, necessitating action binding.
We further scale our model to up to seven actors and train one model that generalizes across 46 games~\cite{agapiou2023meltingpot20} with a unified action space.

Some model-based multi-agent reinforcement learning methods also build a multi-actor world model for policy learning~\cite{wang2026factoredlatentactionworld, wang2025vdfdmultiagentvaluedecomposition, xue2025learning}.
However, these approaches often train small-scale models from scratch~\cite{DreamerV1}.
In contrast, our approach is applicable to large-scale pre-trained video DiT, which proves more scalable.

\heading{Motion control in video diffusion models.}
Several works have studied object trajectory control~\cite{DragNUWA, DragAnything, Tora, SG-I2V, wang2024boximatorgeneratingrichcontrollable, geng2024motionprompting} in generated videos.
They often provide auxiliary control signals such as object masks and bounding boxes along the object trajectory.
Another line of work transfers the motion of a reference video to a newly synthesized one~\cite{SMM, MOFT, pondaven2025ditflow, MotionDirector, VMC, MatchDiffusion, gokmen2025ropecrafttrainingfreemotiontransfer}.
Human motion generation \cite{tevet2022humanmotiondiffusionmodel, zhang2022motiondiffusetextdrivenhumanmotion, jiang2023motiongpthumanmotionforeign} focuses on generating realistic motion sequences or poses from textual descriptions. Some works model sequences of multiple actions within a single motion trajectory \cite{chi2024m2d2mmultimotiongenerationtext, guidedmotion}. However, these approaches mainly generate motion for a single subject at a time and operate in structured pose spaces rather than controlling multiple agents within a visual scene. Game settings involve more abstract actions than pure motion, \eg, interacting with items, firing a beam.
These actions depend on the current state of the subject, such as its current orientation.
Therefore, existing motion control techniques are not suitable for our game environment settings.

\heading{Attribute binding in generative models.}
Attribute binding~\cite{rassin2024linguisticbindingdiffusionmodels, greff2020bindingproblemartificialneural} refers to the challenge of associating specific attributes like text descriptions with corresponding entities within the generated scene. Prior work has studied compositional generation~\cite{zarei2025improvingcompositionalattributebinding, chefer2023attendandexciteattentionbasedsemanticguidance, liu2023compositionalvisualgenerationcomposable, t2icompbenchpp, NeuralAssets} to bind appearance descriptions or spatial relations to different subjects without mixing properties, often implemented with some spatial masking mechanism. Scene graph guided generation~\cite{farshad2023scenegenie, gao2024graphdreamer, liu2025controllable} condition on a structured map of entities, attributes, and spatial relations. Spatio-temporal binding of attributes in video is an even more complex problem that is unsolved~\cite{MinT, LLMGroundedDM, VideoDirectorGPT}, especially when multiple entities are present in the video.

%% file: sec/4_method.tex
\section{Method}

\begin{figure}[!t]
  \centering
  \includegraphics[width=\textwidth]{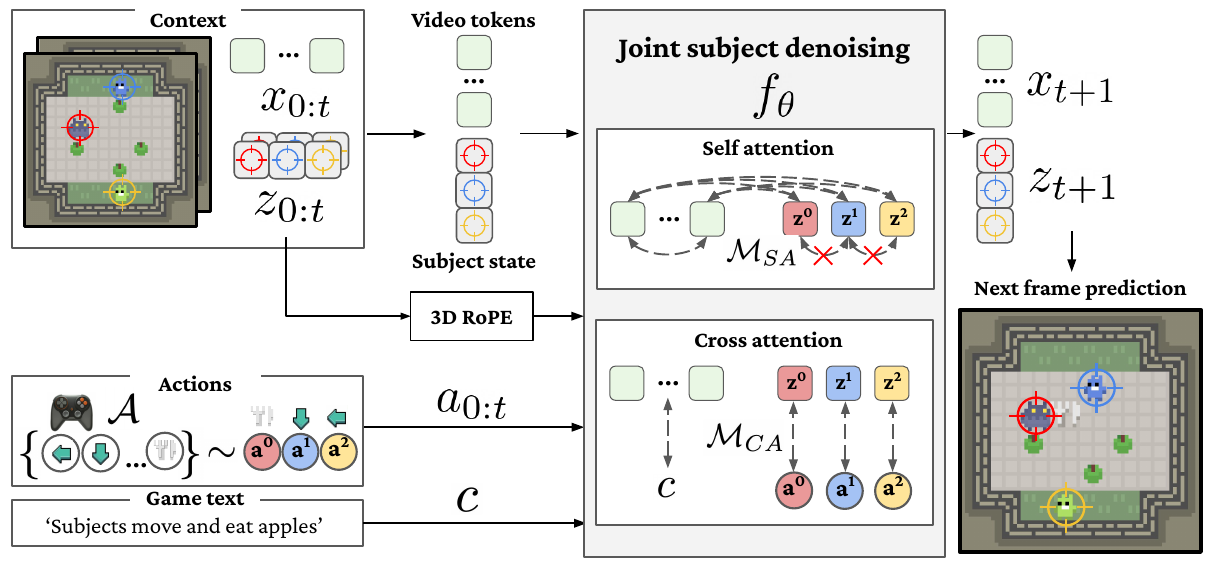}
  \caption{
  \methodname pipeline.
  Given initial video frames $x_{0:t}$ and subject states $z_{0:t}$ as context, we aim to generate the next video frame $x_{t+1}$ conditioned on action inputs $a_{0:t}$ and a text description of the game $c$.
  We concatenate the video and subject state tokens along the sequence dimension and feed them into a diffusion transformer (DiT) for joint denoising.
  Each DiT block first runs self-attention with an attention mask $\mathcal{M}_{SA}$ and 3D RoPE biasing to render subject states to pixels in the video.
  It then runs cross-attention with another attention mask $\mathcal{M}_{CA}$ with explicit subject-action binding to update subject states using action inputs.
  }
  \label{fig:arch}
  \vspace{\figmargin}
\end{figure}

We propose \methodname for multi-subject action-conditioned video generation (\cref{subsec:problem-setup}).
To associate subjects with corresponding actions, we run video diffusion models to jointly denoise video frames and latent tokens capturing the state of each subject (\cref{sec:jointdiff}).
We design an update-and-render paradigm with attention masking (\cref{subsec:generative-game-engine}), and train our model to achieve autoregressive video generation (\cref{sec:autoregresive}).
The overall architecture of \methodname is shown in \cref{fig:arch}.

\subsection{Problem Setup}
\label{subsec:problem-setup}

In this paper, we study controllable video generation for gaming and world-modeling scenarios where multiple entities, referred to as subjects, act within a shared observation space (the same video in this paper). The primary challenge, largely unaddressed in prior single-subject world models, is \textit{action binding}. When multiple controllable subjects are present in a video, the model must correctly associate each control signal (typically an action) with its corresponding subject.

Formally, we define a video as a sequence of frames $x_{0:T}$. We consider a set of different environments $\mathcal{G}=\{g_1, g_2, \dots, g_M\}$. Within a specific environment $g$, we observe $N_g$ controllable subjects $\{S_1,\ldots,S_{N_g}\}$. Our objective is to generate future video frames conditioned on a global text description of the game, several initial context frames, and a sequence of action inputs for each subject.

For simplicity, in this paper we only consider the control spaces with discrete actions. Specifically, at every timestep $t \in \{0,\ldots,T-1\}$, we apply one discrete action per subject, $a_t^i \in \mathcal{A}$, where the action space is fixed across all environments: $\mathcal{A}=\{\mathcal{A}_0,\ldots,\mathcal{A}_{K-1}\}$. Note that the actions here are treated as abstract ``buttons'': the same $\mathcal{A}_k$ can have different effects depending on the environment $g$ and the subject's current state. For example, an ``interact'' command might cause a subject to pick up an object in one game or open a door in another, requiring the model to learn context-dependent dynamics.

Let $a_t \coloneqq \{a_t^i\}_{i=1}^{N_g}$ denote the set of actions applied to each subject at time $t$.
A standard world model $f_\theta$ generates the next frame by sampling from a conditional distribution:
\begin{equation}
    \label{eq:world_model}
    x_{t+1} \sim f_\theta\!\left(x_{t+1} \mid x_{0:t},\, a_{0:t}\right).
\end{equation}
By rolling this process forward for $t=\{t+1,\ldots,T-1\}$, the model produces a controlled trajectory.
In practice, $f_\theta$ is implemented as a conditional video diffusion model.
We assume it can utilize not only the current action $a_t$, but also all previous actions $a_{0:t-1}$ as context.
Therefore, the key question here is: how can the model associate external action signals with subjects that are only implicitly defined within the pixel space?
Beyond their visual appearance, there is no explicit representation of the subjects to which actions can be anchored.
To address this, we propose to augment the world model with additional latent variables representing the subject state.

\subsection{Subject State}
\label{sec:jointdiff}

The main motivation behind our approach is to enable the model $f_\theta$ to disambiguate between different subjects via explicit grounding.
To this end, we introduce an explicit variable ${z_t^i}$ representing the state of subject $S_i$ at time $t$.
To simplify notation, we let $z_t \coloneqq \{z_t^i\}_{i=1}^{N_g}$ denote the set of states for all subjects at a given timestep $t$, and let $z \coloneqq \{z_t\}_{t=0}^T$ represent the state trajectory over the entire sequence.
Consequently, the world model objective in Eq.~\eqref{eq:world_model} can be reformulated as a joint prediction task over $x$ and $z$:
\begin{equation}
    \label{eq:state_world_model}
    x_{t+1}, z_{t+1} \sim f_\theta\!\left(x_{t+1}, z_{t+1} \mid x_{0:t},\, a_{0:t}, z_{0:t}\right).
\end{equation}
To implement this, we extend the standard video DiT architecture (e.g., Wan~\cite{Wan21}) to jointly denoise $x$ and $z$, similar to MMDiT~\cite{SD3}.
Specifically, we concatenate the flattened video tokens $x$ and subject state tokens $z$ along the sequence dimension, and run DiT on the combined sequence of tokens.
To train such a model, we require a concrete definition for the state $z$.
Ideally, this representation should be minimal yet sufficient to disambiguate subjects during complex interactions.
For most game environments, we find that the spatial position of a subject suffices, since two subjects cannot occupy the same position.
Therefore, throughout this paper, we define a subject state as its 2D coordinates $z^i_t = (h^i_t, w^i_t)$.

\begin{figure}[t]
  \centering
  \begin{subfigure}{0.48\linewidth}
    \centering
    \includegraphics[height=4cm]{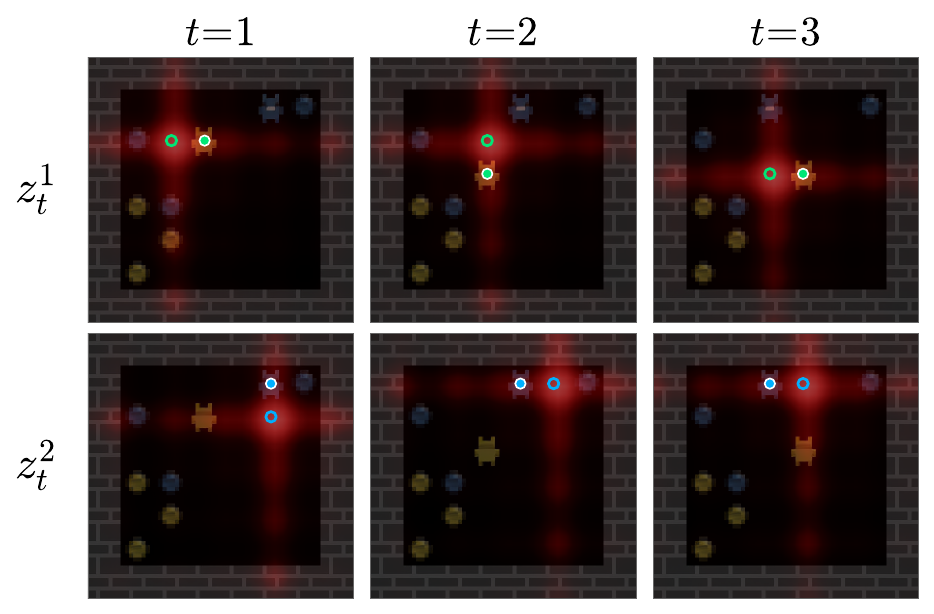}
    \caption{Self-attention subject-video RoPE bias}
    \label{fig:sa_rope}
  \end{subfigure}
  \hfill
  \begin{subfigure}{0.48\linewidth}
    \centering
    \includegraphics[height=4cm]{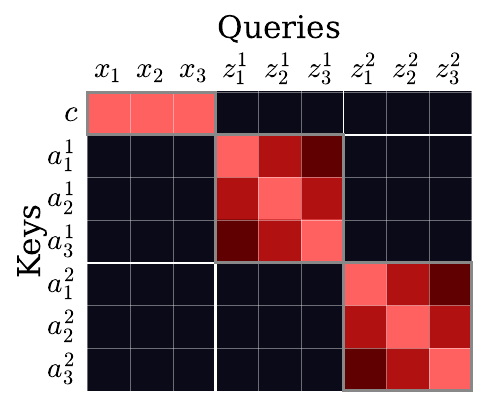}
    \caption{Cross-attention bias between $a_t^i$ and $z_t^i$}
    \label{fig:ca_bias}
  \end{subfigure}
  \caption{
  Attention mechanisms in \methodname DiT.
  (a) In self-attention, we use RoPE to link a subject in a video frame to its state token $z_i^t$.
  We encode the state token with the subject's coordinates in the \emph{previous} timestep, biasing it to attend to video tokens close to the subject.
  (b) In cross-attention, subject $i$'s state tokens $z^i$ is only allowed to attend to its own actions $a^i$, ensuring correct subject-action binding.
  We also allow the text embedding of the environment description $c$ to attend to video tokens $x$.
  }
  \label{fig:short}
  \vspace{\figmargin}
\end{figure}

\subsection{Generative Game Engines} %
\label{subsec:generative-game-engine}

To build an architecture that effectively leverages $z$, we draw an analogy to standard game engines, which decompose the generation of a single frame into two distinct stages: \emph{state update} and \emph{rendering}.
During state update, game engines use actions corresponding to a specific subject to update its state; during rendering, they use the updated subject states to synthesize the video frame.
We implement this paradigm within our video diffusion framework by controlling information flow through masked attention as shown in \cref{fig:short}.

\heading{State update in cross-attention.}
We modify the DiT cross-attention layers to process action inputs and update subject states as shown in \cref{fig:ca_bias}.
Each discrete action $a_t^i$ is first projected to embeddings that match the latent dimension of the model.
We then concatenate the text embedding of the environment description $c$ and all action embeddings to run cross-attention with video and subject state tokens.
We apply a mask $\mathcal{M}_{CA}$ that enforces strict subject-action binding: subject $i$'s state tokens $z^i$ are only allowed to attend to its corresponding action embeddings $a^i$, blocking all other communication between unmatched actions and subject tokens.
To preserve the capacity of the pre-trained model, we restrict the text embedding $c$ to attending solely to the video tokens $x$.

\heading{Rendering in self-attention.}
To render video frames, we apply a mask $\mathcal{M}_{SA}$ in DiT self-attention layers that encourages subject isolation while enabling visual integration.
This mask allows all subject tokens to attend to the video tokens, enabling them to read information from the environment.
Yet, we block subject-to-subject communication to avoid mixing the state between different subjects.
On the other hand, the video tokens attend to all subject tokens, allowing the model to use the state of all subjects for frame synthesis.

\heading{Subject-binding with RoPE bias.}
While the self-attention mask $\mathcal{M}_{SA}$ prevents subject mixing, it does not explicitly link a subject in the video to its state tokens.
Thanks to the use of spatial coordinates as subject states, we can leverage the 3D Rotary Position Embeddings (RoPE) to disambiguate this mapping.
We design a RoPE biasing mechanism within the self-attention layers, where subject tokens are biased to attend to video tokens at the same spatial coordinates.

Specifically, a video token at position $p=(h, w)$ in frame $x_t$ receives a rotation $\mathcal{R}(t,h,w)$.
Given the subject positions from the previous timestep $z_{t-1}^i = (h_{t-1}^i, w_{t-1}^i)$, we apply a rotation $\mathcal{R}(t,h_{t-1}^i, w_{t-1}^i)$ to the subject token $z_t^i$.
We utilize the previous spatial position because the current position $z_t^i$ is being denoised and thus unknown at the current stage.
Still, $z_{t-1}^i$ should be close to the true $z_t^i$, ensuring that video tokens close to the subject are biased to attend to the corresponding subject state token.
Consequently, the model's task is greatly simplified from global disambiguation to local refinement within a single action's radius, as illustrated in \cref{fig:sa_rope}.

\subsection{Training and Inference}
\label{sec:autoregresive}

We train the model on sequences of video tokens and subject state tokens, $(x_{0:t+1}, z_{0:t+1})$, where the context $(x_{0:t}, z_{0:t})$ consists of fully clean ground-truth data, while the targets $x_{t+1}$ and $z_{t+1}$ are noisy. In other words, we train the model $f_\theta$ to denoise $x_{t+1}$ and $z_{t+1}$ conditioned on the clean context $(x_{0:t}, z_{0:t})$. To support variable-length contexts, we use sequences up to length $T$ and pad the positions from $t+2$ to $T$ with fully noisy frames.

During inference, we assume that the initial frame $x_0$ and the initial subject positions $\{{z_0^i}\}_{i=1}^{N_g}$ are provided. Note that without this initialization, distinguishing between subjects would be impossible in many environments where entities share an identical appearance. We then perform an autoregressive rollout: given the previously generated context $(x_0||\hat{x}_{1:t}, z_0^i||\hat{z}^i_{1:t})$, where $\hat{x}$ and $\hat{z}$ are previous model predictions, we predict the subsequent video frame $x_{t+1}$ and state variables $z_{t+1}$. When $t > T$, we drop the oldest frames from the context, ensuring the context window size never exceeds $T-1$.

%% file: sec/5_experiments.tex
\section{Experiments}

\subsection{Setup}

\heading{Implementation details.}
We fine-tune Wan2.1-1.3B~\cite{Wan21}, an open-source text-to-video (T2V) DiT with the training setup outlined in \cref{sec:autoregresive}.
To speed up model convergence, we first pretrain the model on raw game videos to adapt to the autoregressive setup.
For this stage, we only provide text conditioning, which is a simple description of the game, without action control, subject state, or any architecture modification.
We train on all games ($|\mathcal{G}|=46$) for 22.5k steps with a batch size of 64.
Then, we add action control and fine-tune the full \methodname model for 65k steps with the same batch size.
We train our method with $T=5$ timesteps, corresponding to 4 unique actions per rollout.
We use Adam~\cite{Adam} to optimize the flow matching loss~\cite{FlowMatching, RFSampler} computed on $x_{t+1}^i$ and $z_{t+1}^i$.
We consider a maximum of 7 subjects in a scene and use zeros for padded subject states.
At inference time, we sample for 20 steps with a timestep shift of 5.0~\cite{SD3}.
We implement attention masking with FlexAttention~\cite{FlexAttention}, only adding $N \times T$ subject tokens per subject, which brings minimal overhead compared to the number of video tokens.
In practice, we only add a 6\% overhead for 7 players for $T=5$ (40 extra tokens).
Note that this is significantly more efficient than other methods~\cite{enigma2025multiverse, solaris2026} that generate multiple views of the scene for each subject.

\heading{Datasets and action space.}
We train \methodname on gameplay videos of 46 different 2D games from the Melting Pot benchmark~\cite{agapiou2023meltingpot20}.
These games feature character sprites moving on a tile grid with different environments and static objects.
Some characters may look identical, making action binding more challenging.
\methodname is trained with an action set $\mathcal{A}$ including 25 different actions.
Notably, some actions are only present in specific games.
There are 7 base actions shared by all games, which we categorize in 4 different groups: \textit{Idle action} (staying still), \textit{Move actions}: \{forward, backward, strafe left, strafe right\}, \textit{Turn actions}: \{turn left, turn right\} and \textit{Interact}.
The ``Interact'' action involves game-specific interaction with the environment, such as highlighting tiles in front of it to fire a beam.
Importantly, subject movements are relative to the subject's orientation.
For example, the ``forward'' action moves a sprite one tile forward in the direction it is facing.
This is particularly interesting in our setup as subject orientations can only be inferred from raw video frames; hence, the model has to comprehend visual information to get the correct action outcome, rather than learning absolute directional instructions like ``move up''.
We generate game-specific text descriptions using an LLM from the Melting Pot paper and videos of each game (full details in Supplementary).
For training, we gather 2,000 videos per game at a resolution of 512$\times$512.
We generate diverse rollouts for each game by executing either random actions or pretrained policies at each step $t$.
For evaluation, we collect 230 rollouts (5 per game).
As completely random actions result in players staying close to their initial positions, we enforced $a_t^i \sim \{Move, Turn\}$ for $t<2$ to leave its initial state, and at least one subject is assigned an ``Interact'' action for $t \geq 2$.

\heading{Baselines.}
Since there is no open-source method capable of controlling more than two subjects, we design several baselines.
To demonstrate the challenge of action binding from text prompts, we propose a naive \textit{Zero-shot I2V} baseline where we simply prompt a video model with a textual description of the initial subject positions and the actions to execute.
We use the Wan2.1-14B I2V model~\cite{Wan21}, which is much larger than the 1.3B model \methodname fine-tunes from.
We also study such \textit{Text-Action} control on our autoregressive setup, using the same backbone as \methodname, but trained only using textual conditioning.
Finally, we take the autoregressive model pretrained on game videos without action control (termed \textit{Pretrained AR}) as a random action lower bound.

\heading{Evaluation.}
The effect of an action on a subject depends on its state and the environment, e.g., position/orientation and occlusion.
To evaluate action binding, direct visual comparison with ground-truth videos is unreliable as the correct action may be performed even if the subject's state has drifted. 
We propose a Movement Accuracy metric (MA) that extracts each subject's movement (left, right, up, down, or still) from consecutive video frames and compares it with the input action.
To do so, we train simple subject detectors that achieve near 100\% accuracy on raw video frames.
To measure the effect of context-dependent ``Interact'' actions, we introduce an Effect Accuracy metric (EA) that extracts a 3$\times$3 tile patch centered on the ground truth player position and evaluate dissimilarity with the previous frame with SSIM.
We empirically set a threshold patch SSIM $\geq 0.85$ to detect executed actions.
We also introduce two simple metrics to monitor subjects: Subject Preservation (SP) and Detection Rate (DR).
SP measures how many subjects are retained at the end of the generation.
DR measures the percentage of steps where the subject's position aligns with ground-truth.
Please refer to Supplementary for more details on metric implementations.
For completeness, we also report standard visual quality metrics between ground-truth and generated videos, including PSNR, LPIPS~\cite{LPIPS}, and FVD~\cite{FVD}.
Finally, as our model also predicts subject coordinates, we compute $z_t$ error as the L2 distance between ground-truth and predicted positions.

\begin{table}[!t]
  \caption{Baseline comparison on action binding metrics and visual fidelity.
  \methodname is the only method able to associate actions to the correct characters (MA), preserve the subject appearance (SP), and execute rollouts with correct subject trajectories (DR).
  We also compute fidelity metrics, LPIPS, PSNR, and FVD, which show that videos generated by \methodname are consistent with ground-truth.
  }
  \label{tab:metrics}
  \centering
  \setlength{\tabcolsep}{6pt}
  \begin{tabular}{@{}lcccccc@{}}
    \toprule
    \multirow{2}{*}[-.3em]{\textbf{Method}} & \multicolumn{3}{c}{Action Binding} & \multicolumn{3}{c}{Visual Quality} \\
    \cmidrule(lr){2-4}
    \cmidrule(lr){5-7}
    & \textbf{MA} $\uparrow$ & \textbf{SP} $\uparrow$ & \textbf{DR} $\uparrow$ & \textbf{LPIPS} $\downarrow$ & \textbf{PSNR} $\uparrow$ & \textbf{FVD} $\downarrow$ \\
    \midrule
    Pretrained AR & 0.065 & 0.571 & 0.330 & 0.0494 & 26.98 & 59.30 \\
    Zero-shot I2V & 0.027 & 0.422 & 0.440 & 0.0906 & 23.17 & 281.55 \\
    Text-Action & 0.158 & 0.668 & 0.433 & 0.0353 & 29.14 & 56.74 \\
    Ours (\methodname) & \textbf{0.779} & \textbf{0.903} & \textbf{0.886} & \textbf{0.0102} & \textbf{36.35} & \textbf{17.16} \\
    \bottomrule
  \end{tabular}
\end{table}

\begin{table}[!t]
  \caption{We evaluate the Effect Accuracy of our generated videos across different action types.
  \methodname achieves the highest accuracy on ``Interact'' actions compared to other methods, indicating that it most often creates the desired visual effect on neighboring tiles across different games.
  }
  \label{tab:action_effects}
  \centering
  \setlength{\tabcolsep}{8pt}
  \begin{tabular}{@{}lccccc@{}}
    \toprule
    \multirow{2}{*}[-.3em]{\textbf{Method}} & \multicolumn{5}{c}{Effect Accuracy $\uparrow$}\\
    \cmidrule(lr){2-6}
    & Idle & Move & Turn & Interact & Overall \\
    \midrule
    Pretrained AR & 0.399 & 0.414 & 0.462 & 0.346 & 0.412 \\
    Zero-shot I2V & 0.000 & 0.101 & 0.150 & 0.000 & 0.126 \\
    Text-Action & 0.357 & 0.420 & 0.553 & 0.326 & 0.433 \\
    Ours (\methodname) & \textbf{0.899} & \textbf{0.867} & \textbf{0.914} & \textbf{0.774} & \textbf{0.861} \\
    \bottomrule
  \end{tabular}
\end{table}

\subsection{Action binding evaluation}

\heading{Quantitative evaluation.}
We compare action binding performance for \methodname and baselines in \cref{tab:metrics}. %
First, we notice that baselines dramatically fail in movement accuracy, for which the best baseline (Text-Action) achieves 0.158.
\methodname instead yields \textbf{0.779}, showcasing that our contributions are fundamental to correctly bind actions to specific subjects.
Interestingly, we noticed that baselines tend to remove the subject from the scene, as highlighted from our improved SP (\textbf{0.903}) with respect to the closest baseline (0.668).
Finally, we report improvements in DR, showing that subjects are rendered in the correct position most of the time (\textbf{0.886}).
The trained methods demonstrate consistency with ground-truth, with ours achieving the highest alignment due to improved action following as shown by the visual fidelity metrics. 
We also break down the effect accuracy in each category in \cref{tab:action_effects}.
\methodname works the best across all types of actions. Especially on the context-dependent ``Interact'' action, our model more than doubles the success rate of existing baselines.

\begin{figure}[!t]
  \centering
  \includegraphics[width=\linewidth]{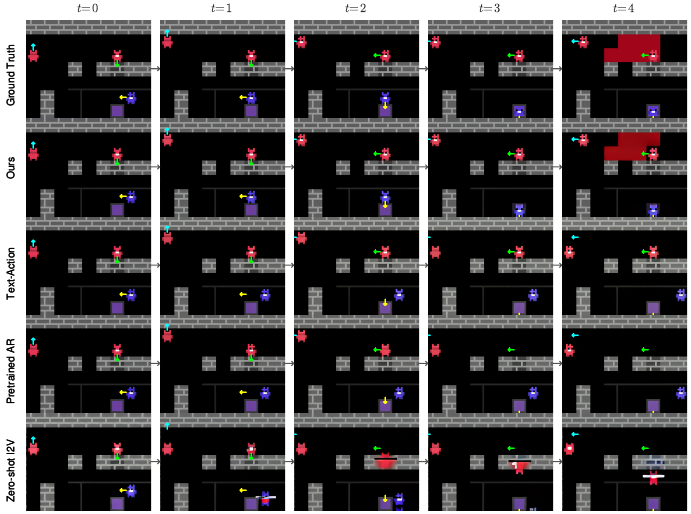}
  \caption{Qualitative comparison with baselines. We display ground truth subject positions and orientations at each step with an arrow for each subject, which reflects the ground truth subjects (notice that the arrow is consistent in all methods). Our method is the only one able to follow the ground truth actions with appropriate action binding.}
  \label{fig:qual_baselines}
  \vspace{\figmargin}
\end{figure}

\begin{figure}[!t]
  \centering
  \includegraphics[width=0.6\linewidth]{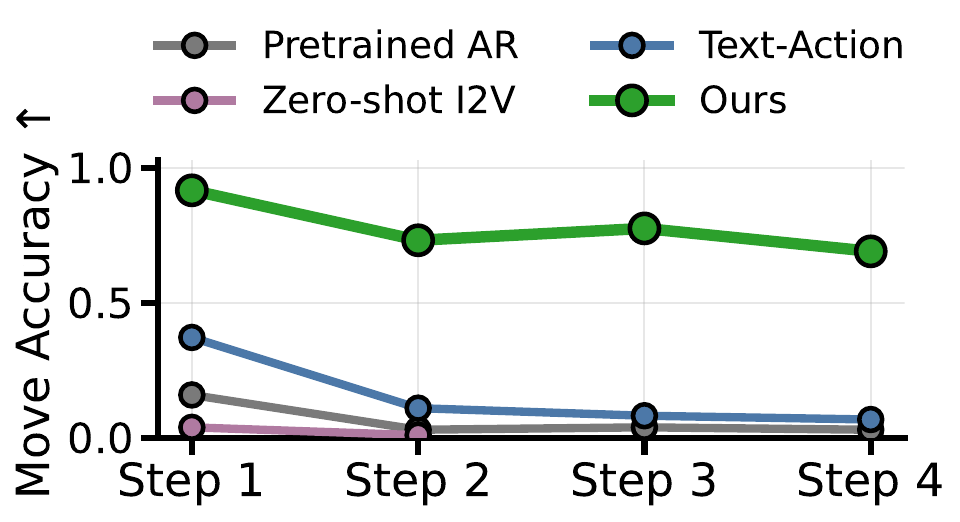}
  \caption{Movement accuracy (MA) over autoregressive steps. \methodname maintains stable action binding across multiple rollout steps, whereas baselines degrade over time and get close to 0.}
  \label{fig:act-stability}
  \vspace{\figmargin}
\end{figure}

\heading{Autoregressive stability.}
In \cref{fig:act-stability}, we measure the stability of our autoregressive inference.
We plot, for different autoregressive steps, the movement accuracy of \methodname and baselines.
While some baselines, such as Text-Action, yield suboptimal yet acceptable results on the first step, it quickly degrades in subsequent steps.
Conversely, thanks to the joint modeling of subject states, \methodname allows for a relatively stable autoregressive inference, showing that our action binding performance remains consistent over time.

\begin{figure}[!t]
  \centering
  \includegraphics[width=\linewidth]{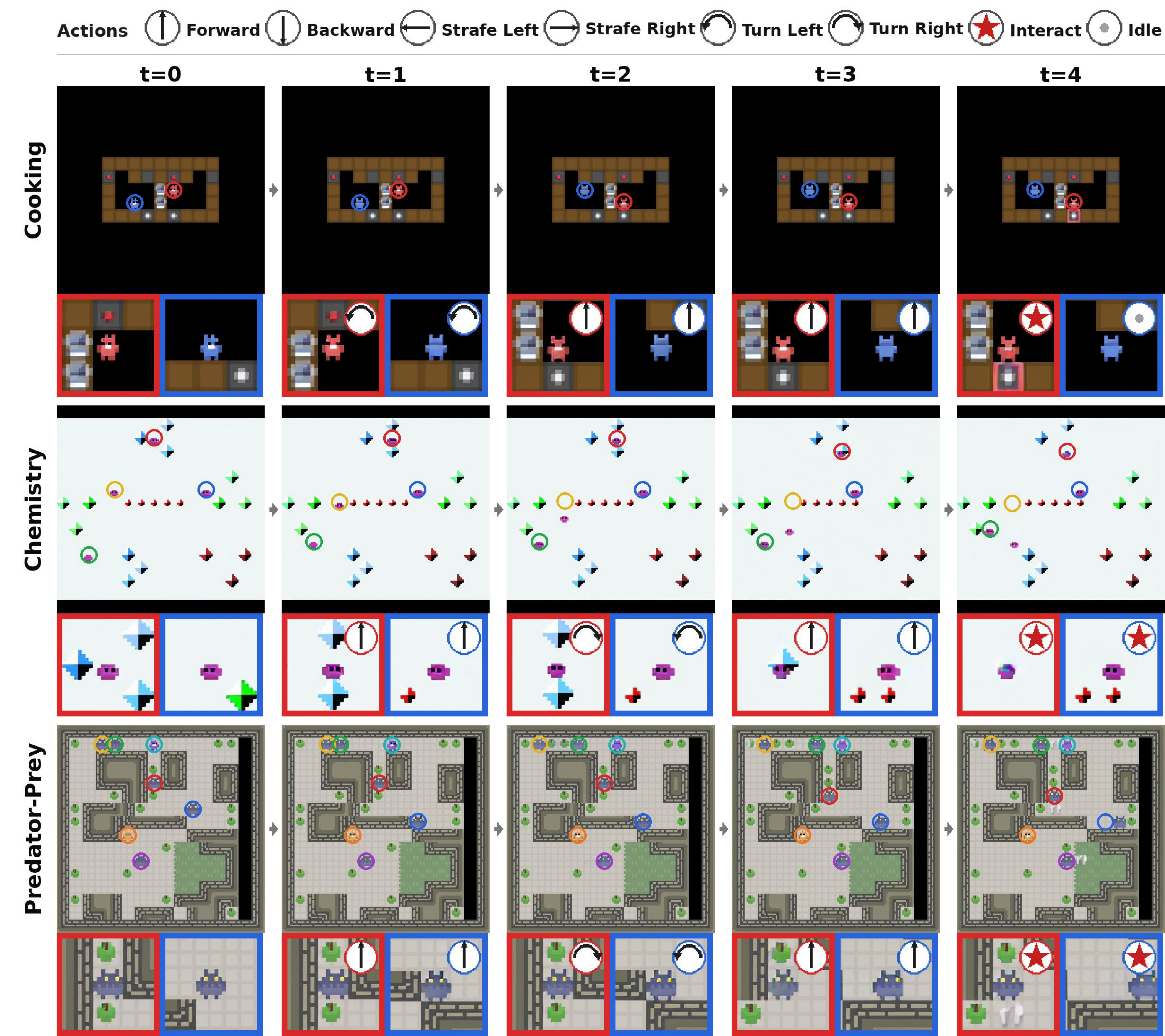}
  \caption{Results of \methodname with $x_t$ and $z_t$ (circle) predictions for 2-, 4-, and 7-player games with a diversity of subjects. The full frames are generated and we show a centred crop around 2 subjects for visualization purposes with action annotations on the \emph{resulting} frame. Note that actions are \emph{relative to orientation}, so the subject moving forward is in the direction of where it is facing.
  }
  \label{fig:qual}
  \vspace{\figmargin}
\end{figure}

\heading{Qualitative comparison.} 
In \cref{fig:qual_baselines}, we compare \methodname with baselines on a sample from the Paintball game (more results in Supplementary).
In this game, subjects can move around, turn, and shoot a colored beam with the ``Interact'' action.
In the figure, we use small arrows to denote the ground-truth position and orientation of all subjects at a given step.
Compared to baselines, \methodname is the only method that successfully binds all actions to subjects.
Our rendered frames are similar to the ground-truth, with arrows overlaid on subjects, and we successfully bind the ``Interact'' action in the last step, as a red beam appears in front of the subject.
Conversely, the Text-Action baseline fails whenever subjects move away from their initial positions, as it is unable to differentiate distinct subjects.
The zero-shot I2V approach fails drastically at assigning separate actions to different subjects and applies a continuous downwards movement to detected sprites.
It also fails to maintain the appearance of subjects.
This aligns with our reported metrics, and showcases the importance of our problem setup as a common failure mode of video diffusion models.

\heading{Qualitative results.} In \cref{fig:qual}, we test our method on a range of games including Cooking, Chemistry, and Predator-Prey, which have 2, 4, and 7 subjects being controlled, respectively. We plot a zoomed-in view of two subjects in each frame to aid visibility, however all subjects are indeed being controlled in a single scene. Coordinate predictions $z_t$ are plotted as colored circles. Our position control can successfully control identical-looking subjects as in the Chemistry game. Our single learned action space can also apply the same actions to diverse appearances in unique backgrounds as seen in the bottom two rows. Even if the coordinate prediction is slightly off, action binding may still be successful if it is enough to spatially disambiguate different subjects.

\begin{table}[!t]
  \caption{
  Ablations on the 4-action Coins game. %
  The full \methodname model achieves the most accurate localization of players indicated by the highest movement accuracy (MA) and the lowest $z_t$ error.
  Our $\mathcal{M}_\text{CA}$ is important for consistent subject preservation according to the subject appearance (SP) and detection rate (DR).
  }
  \label{tab:ablations}
  \centering
  \setlength{\tabcolsep}{8pt}
  \begin{tabular}{@{}lcccc@{}}
    \toprule
    Method & \textbf{MA} $\uparrow$ & \textbf{$z_t$ Error} $\downarrow$ & \textbf{SP} $\uparrow$ & \textbf{DR} $\uparrow$\\
    \midrule
    w/o $\mathcal{M}_{SA}$  & 0.580 & 0.108 & \textbf{1.00} & 0.716 \\
    w/o $\mathcal{M}_{CA}$  & 0.052 & 0.215 & 0.88 & 0.307 \\
    Frame-wise $\mathcal{M}_{CA}$  & 0.052 & 0.240 & 0.80 & 0.293 \\
    No RoPE in SA  & 0.032 & 0.291 & \textbf{1.00} & 0.278 \\
    Ours (\methodname) & \textbf{0.872} & \textbf{0.072} & \textbf{1.00} & \textbf{0.913} \\
    \bottomrule
  \end{tabular}
\end{table}

\begin{figure}[!t]
  \centering
  \includegraphics[width=\linewidth]{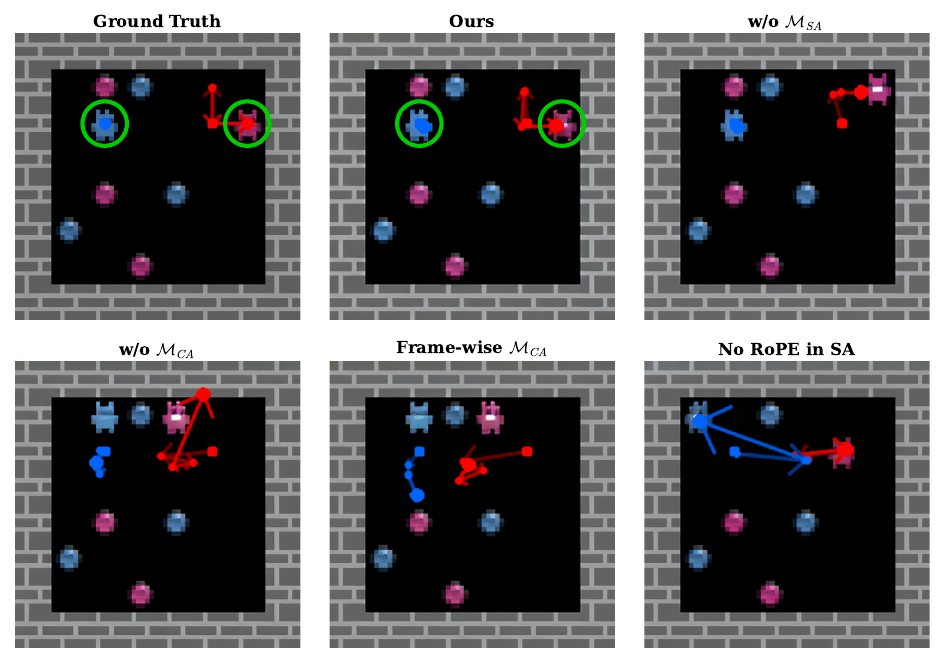}
  \caption{Qualitative comparison of ablations run on the same initial frame and action trajectory. The blue subject is still, while the purple one moves and returns to its previous position. The $z_t$ trajectory is plotted for each subject. If we remove our introduced components, we lose the possibility to control characters.
  }
  \label{fig:ablations}
  \vspace{\figmargin}
\end{figure}

\subsection{Ablation studies}

We conduct ablation studies in a smaller-scale setup due to excessive training costs.
Here, we collect 2.5k samples from one of the game environments called Coins at a resolution of 256$\times$256.
This game involves two players walking around and picking up coins that are randomly scattered around a map.
All ablations are initialized from the same autoregressive pretrained checkpoint and fine-tuned with action control for 45k steps.
In this setup, our model achieves \textbf{87.2\%} movement accuracy (MA) and maintains a subject detection rate (DR) at ground truth positions of \textbf{91.3\%}.

We list results in \cref{tab:ablations} and qualitatively compare the final generated frame and subject trajectories $z_t$ in \cref{fig:ablations}.
We first study the importance of the attention masks in \methodname.
By removing the self-attention mask $\mathcal{M}_\text{SA}$, subject tokens are allowed to attend to each other, causing information leak between subjects and ultimately reducing MA to 58\%.
We also tried removing the cross-attention mask $\mathcal{M}_\text{CA}$, which results in a drop of MA to 5.2\%.
Evidently, the presence of $\mathcal{M}_\text{CA}$ ensures the assignment of actions to the correct subject state tokens, and thus is a fundamental component for correct action binding.
Similarly, adding a frame-wise $\mathcal{M}_\text{CA}$ that only allows each subject token to attend to the action at its own frame, rather than all frames, reduces performance drastically.
Attending to the full action sequence in $z_t$ may improve action following.
Finally, removing the RoPE bias in self-attention prevents any action binding as the subject tokens do not attend to the correct video tokens (MA drops to 3.2\%).
In general, we notice that the $z_t$ error and detection rate closely align with MA, meaning that action binding relies on accurate localization of subjects.

%% file: sec/6_conclusions.tex
\section{Conclusion}

In this paper, we present \methodname, a framework for resolving the action binding challenge in multi-player game simulations.
By introducing subject state tokens that are jointly denoised with video latents, we successfully associate unique, controllable identifiers with individual subjects across 46 diverse environments.
Through attention masking and spatial RoPE biasing, \methodname disentangles subject update and frame rendering, enabling precise control of up to seven subjects simultaneously.
Our results demonstrate that explicit subject grounding significantly improves action-following accuracy and identity preservation in the autoregressive video diffusion setup.
We believe our study addresses a fundamental bottleneck in action-conditioned video generation, and we hope it can inspire future research on robust, multi-subject video world models.

%% file: sec/7_acknowledgements.tex
\section*{Acknowledgements}

Alexander Pondaven is generously supported by a Snap studentship and the EPSRC Centre for Doctoral Training in Autonomous Intelligent Machines and Systems [EP/S024050/1].

%% file: sec/X_suppl.tex
\clearpage
\setcounter{page}{1}
\setcounter{section}{0}
\setcounter{table}{4}
\maketitlesupplementary
\appendix

We encourage readers to see video results in the website. Note that actions in the MeltingPot dataset are relative to the orientation of subjects, so the forward action `F' would take one step in the direction the subject is facing. The website includes ActionParty generations and baseline comparisons for each game. We also trained ActionParty with a $T=20$ context window with 20 unique actions and show generation works as in the $T=5$ setup. Finally, we include ablation videos, long horizon generation and generalization to different player counts. In this section, we provide extra details from the main report. We first provide more fine-grained metrics for different subject counts in \cref{sec-supp:persubjectmetrics}. Then, we provide more implementation (\cref{sec-supp:implementation}) and dataset (\cref{sec-supp:dataset}) details. Finally, we discuss some exploration of generalization capabilities in \cref{sec-supp:generalization} and the limitations of our method in \cref{sec-supp:limitations}.

\section{Fine-grained metrics} \label{sec-supp:persubjectmetrics}

We split our metrics into performance over games with different subject counts with the following categories: Sparse (2-3), Dense (4-5) and Crowded (6-7). We have 17, 24, and 5 games in each of these categories respectively. Some games may inherently be more difficult to model, and there may be more variability over fewer games. We show the fine-grained metrics for Tab.~1 in \cref{tab:metrics-by-subjects} and Tab.~2 in \cref{tab:effect-accuracy-by-subjects}. In general, we see that our method maintains consistent action following and visual quality over different player count setups, while baselines are not able to follow actions even when there are just 2-3 subjects being controlled.

\begin{table}[!t]
  \caption{Metrics split by number of subjects per game. All methods degrade with more subjects, but \methodname remains the most robust across all metrics.}
  \label{tab:metrics-by-subjects}
  \centering
  \setlength{\tabcolsep}{5pt}
  \small
  \begin{tabular}{@{}llccccccc@{}}
    \toprule
    \multirow{2}{*}[-.3em]{\textbf{Method}} & \multirow{2}{*}[-.3em]{\textbf{\# Players}} & \multicolumn{3}{c}{Action Binding} & \multicolumn{3}{c}{Visual Quality} \\
    \cmidrule(lr){3-5}
    \cmidrule(lr){6-8}
    & & \textbf{MA} $\uparrow$ & \textbf{SP} $\uparrow$ & \textbf{DR} $\uparrow$ & \textbf{LPIPS} $\downarrow$ & \textbf{PSNR} $\uparrow$ & \textbf{FVD} $\downarrow$ \\
    \midrule
    \multirow{3}{*}{Pretrained AR}
      & Sparse  & 0.079 & 0.649 & 0.366 & 0.0281 & 29.27 & 53.31 \\
      & Dense   & 0.064 & 0.534 & 0.317 & 0.0597 & 25.71 & 84.53 \\
      & Crowded & 0.051 & 0.580 & 0.331 & 0.0673 & 25.33 & 293.25 \\
    \midrule
    \multirow{3}{*}{Zero-shot I2V}
      & Sparse  & 0.031 & 0.485 & 0.435 & 0.0605 & 24.57 & 229.20 \\
      & Dense   & 0.018 & 0.385 & 0.418 & 0.1055 & 22.32 & 427.58 \\
      & Crowded & 0.035 & 0.463 & 0.441 & 0.1218 & 22.47 & 678.30 \\
    \midrule
    \multirow{3}{*}{Text-Action}
      & Sparse  & 0.196 & 0.830 & 0.499 & 0.0173 & 32.40 & 49.66 \\
      & Dense   & 0.155 & 0.643 & 0.414 & 0.0430 & 27.47 & 81.06 \\
      & Crowded & 0.125 & 0.548 & 0.414 & 0.0542 & 26.46 & 245.87 \\
    \midrule
    \multirow{3}{*}{Ours}
      & Sparse  & \underline{0.781} & \underline{0.878} & \textbf{0.887} & \textbf{0.0063} & \textbf{38.56} & \textbf{16.68} \\
      & Dense   & 0.772 & \textbf{0.943} & \underline{0.886} & \underline{0.0111} & \underline{35.47} & \underline{25.97} \\
      & Crowded & \textbf{0.797} & 0.822 & 0.884 & 0.0166 & 33.57 & 46.14 \\
    \bottomrule
  \end{tabular}
\end{table}

\begin{table}[!t]
  \caption{Effect Accuracy split by scene density. \methodname maintains high accuracy across all action types even in Crowded scenes, while baselines degrade significantly. Zero-shot I2V produces only a single next-frame prediction, so Idle and Interact actions (which occur at later steps in validation set) are not evaluated and reported as 0.}
  \label{tab:effect-accuracy-by-subjects}
  \centering
  \setlength{\tabcolsep}{6pt}
  \small
  \begin{tabular}{@{}llccccc@{}}
    \toprule
    \multirow{2}{*}[-.3em]{\textbf{Method}} & \multirow{2}{*}[-.3em]{\textbf{\# Players}} & \multicolumn{5}{c}{Effect Accuracy $\uparrow$} \\
    \cmidrule(lr){3-7}
    & & Idle & Move & Turn & Interact & Overall \\
    \midrule
    \multirow{3}{*}{Pretrained AR}
      & Sparse  & 0.512 & 0.438 & 0.488 & 0.411 & 0.451 \\
      & Dense   & 0.379 & 0.461 & 0.481 & 0.368 & 0.443 \\
      & Crowded & 0.300 & 0.243 & 0.374 & 0.221 & 0.276 \\
    \midrule
    \multirow{3}{*}{Zero-shot I2V}
      & Sparse  & 0.000 & 0.122 & 0.211 & 0.000 & 0.167 \\
      & Dense   & 0.000 & 0.119 & 0.155 & 0.000 & 0.137 \\
      & Crowded & 0.000 & 0.029 & 0.071 & 0.000 & 0.050 \\
    \midrule
    \multirow{3}{*}{Text-Action}
      & Sparse  & 0.419 & 0.460 & 0.662 & 0.370 & 0.497 \\
      & Dense   & 0.368 & 0.454 & 0.563 & 0.343 & 0.456 \\
      & Crowded & 0.233 & 0.275 & 0.401 & 0.234 & 0.299 \\
    \midrule
    \multirow{3}{*}{Ours}
      & Sparse  & \underline{0.907} & \underline{0.864} & \textbf{0.932} & \textbf{0.870} & \textbf{0.887} \\
      & Dense   & \textbf{0.916} & \textbf{0.879} & \underline{0.914} & 0.743 & \underline{0.861} \\
      & Crowded & 0.833 & 0.832 & 0.893 & \underline{0.773} & 0.835 \\
    \bottomrule
  \end{tabular}
\end{table}

\section{Implementation details} \label{sec-supp:implementation}

\paragraph{Details on state token $z_t$}
We need to adapt the 2D coordinates $z_t$ to pass them into the DiT. We normalize coordinates (re-centre and scale by a factor of 4.0) such that they have similar statistics to VAE-encoded video latents. We introduce coordinate encoder and decoder as linear projections, $W_{enc} \in \mathcal{R}^{2\times1536}$ and $W_{dec} \in \mathcal{R}^{1536\times2}$ respectively. This maps 2D coordinates to the inner dimension of the video DiT, allowing subject tokens to be properly concatenated to video tokens, and then decoded back into coordinate space to compare against ground truth for the flow matching loss computation. %

\paragraph{Metrics} The dataset has 7 base movement actions and 18 possible ``Interact'' actions that have different effects depending on the game (25 max actions). Not all games use all actions -- 25 of the games only use 1 ``Interact'' action (8 total actions). For visualization and metrics, we group ``Interact'' actions into one category. In order to measure action accuracy, we introduce a linear detection and classification model to get subject position and orientation. Specifically, to train these linear models, we collect tile images at subject ground truth positions from the dataset and use other tiles as negative data to achieve ~100\% accuracy. Note that comparing to ground truth to evaluate effect performance is not always accurate (\eg the ground truth might have a wall in front of the player and not highlight any blocks, while the generation might not have a wall), however it always punishes drifting from ground truth and lack of action following in the video as a whole, which makes it a suitable metric for overall action binding performance.

\section{Melting Pot details} \label{sec-supp:dataset}

\paragraph{Adapted Melting Pot} The original set of games in the Melting Pot suite has 49 games at a large range of resolutions and scales. For video training, we want a fixed resolution for stability. The Wan~\cite{Wan21} DiT applies VAE encoding, which does spatial downsampling by a factor of 8 and we require good reconstruction for proper action following. We find that player tiles need to be at a pixel scale of 32$\times$32 to properly distinguish player orientation after reconstruction. In order to fit the entire game into a 512$\times$512 resolution with 32$\times$32 pixels per tile, we need to adapt each environment map to be a 16$\times$16 grid. Each environment map is expressed as an ASCII text representation in the Lua-based backend of the Melting Pot library~\cite{agapiou2023meltingpot20}. We use an LLM in order to adapt as many game maps as possible to fit in a 16$\times$16 grid. We test playability of each generated map by rolling out with trained policies to ensure players still receive rewards.

\paragraph{Game and action description format}
We condition the DiT with text prompts describing each game setup. All the prompts are in the website and we display some examples in \cref{tab:game_prompts}.
The \textit{Text-Action} and \textit{Zero-shot I2V} baselines both provide subject action descriptions and positions in the text prompt. Initial positions are provided in the format: `player 1 starts at (col, row). player 2 starts at (col, row)', where we use tile grid coordinates \eg (1,4). The per-frame actions are in the format: `In frame N, player 1 moves forward. player 2 moves backwards.' The Interact action are given game-specific action names such as `fires beam', `interacts', and `fires paintball'. The initial position and per-frame action descriptions up to the frame being denoised are appended to the game text description. This prompt provides the exact same information provided to our method.

\begin{table}[!t]
  \caption{Example game description prompts used for text conditioning. Each prompt describes the visual content of the environment. Full prompts for all 46 games are provided in the supplementary materials.}
  \label{tab:game_prompts}
  \centering
  \small
  \setlength{\tabcolsep}{5pt}
  \begin{tabular}{@{}p{2.2cm}p{\dimexpr\linewidth-2.2cm-2\tabcolsep-2\arrayrulewidth}@{}}
    \toprule
    \textbf{Game} & \textbf{Prompt} \\
    \midrule
    Coins & Two player sprites move around a black-floored room surrounded by gray brick walls. Small pink and yellow coin sprites periodically appear at random positions on the floor. Players walk over coins to pick them up and coins disappear on contact. \\
    \addlinespace
    Clean up & Player sprites move across a map split into a sandy upper area and a green lower area with a blue-green polluted river section between them. Gray brick walls surround the map. Pink apple sprites appear on the green grass. Players fire cleaning beams at the river pollution. \\
    \addlinespace
    Chemistry & Player sprites move across a white open field scattered with colorful arrow-shaped molecule sprites in blue, green, red, and cyan. Players walk into molecules to pick them up and carry them. Molecules near each other react and transform into different colored molecules. \\
    \addlinespace
    Collaborative cooking & Two player sprites move in a very small kitchen made of brown counter tiles with minimal dark floor space. Red tomato sprites, gray cooking pots, and white plate sprites sit on counters. Players squeeze past each other carrying items between stations. \\
    \addlinespace
    Paintball & Two teams of player sprites move through a gray-walled arena bordered by purple indicator tiles. Players fire beams that paint the floor red or blue in large patches. Small flag sprites sit at each team's base. Players pick up the opposing flag and carry it back. \\
    \bottomrule
  \end{tabular}
\end{table}

\section{Generalization} \label{sec-supp:generalization}
\paragraph{Long horizon} We can autoregressively extend generations using a sliding window that is the size of the context length (see Long Horizon Generation section in the website). We use our model trained on 4 rollout steps to generate videos for 20 steps and show consistent predictions. Near the end, the predicted coordinates appear to drift from the subject positions due to error accumulation.

\paragraph{Player count} We show that our method can generalize to unseen player counts for the same scene (see Variable Player Count section in the website). Specifically, the Coins game was only trained with two subjects in the scene and we show we can generate rollouts with 1-8 subjects present. Note that none of the games had 1 or 8 players in the scene. For more subjects, it does appear to leak ``Interact'' action effects from the wrong games.

\section{Limitations} \label{sec-supp:limitations}
Our predicted position $z_t$ may not completely align with the subject after a wrong move, but our smooth position biasing allows actions to still bind to correct subjects. Some subjects can disappear, which prevents further interaction with the scene. Our model is also not fully real-time yet, however with techniques such as distillation and diffusion forcing~\cite{CausVid, chen2024diffusionforcingnexttokenprediction, SelfForcing}, the model can be made more interactive. The main focus of this work is on action binding. We test our method in a 2D game setup as it aligns well with the task of binding actions to multiple subjects within the same scene. Further work may be done to extend this to other game environments like 3D scenes where subjects are partially visible~\cite{solaris2026}.

%% file: main.bib
@String(PAMI  = {IEEE Trans. Pattern Anal. Mach. Intell.})

@String(CVPR  = {IEEE Conf. Comput. Vis. Pattern Recog.})

@String(ICCV  = {Int. Conf. Comput. Vis.})

@String(ECCV  = {Eur. Conf. Comput. Vis.})

@String(NeurIPS = {Adv. Neural Inform. Process. Syst.})

@String(ICML  = {Int. Conf. Mach. Learn.})

@String(ICLR  = {Int. Conf. Learn. Represent.})

@String(TOG   = {ACM Trans. Graph.})

@String(PAMI  = {IEEE TPAMI})

@String(CVPR  = {CVPR})

@String(ICCV  = {ICCV})

@String(ECCV  = {ECCV})

@String(NeurIPS = {NeurIPS})

@String(ICML  = {ICML})

@String(ICLR  = {ICLR})

@String(TOG   = {ACM TOG})

@inproceedings{liu2025controllable,
  title={Controllable 3D outdoor scene generation via scene graphs},
  author={Liu, Yuheng and Li, Xinke and Zhang, Yuning and Qi, Lu and Li, Xin and Wang, Wenping and Li, Chongshou and Li, Xueting and Yang, Ming-Hsuan},
  booktitle={ICCV},
  year={2025}
}

@inproceedings{gao2024graphdreamer,
  title={Graphdreamer: Compositional 3d scene synthesis from scene graphs},
  author={Gao, Gege and Liu, Weiyang and Chen, Anpei and Geiger, Andreas and Sch{\"o}lkopf, Bernhard},
  booktitle={CVPR},
  year={2024}
}

@inproceedings{farshad2023scenegenie,
  title={Scenegenie: Scene graph guided diffusion models for image synthesis},
  author={Farshad, Azade and Yeganeh, Yousef and Chi, Yu and Shen, Chengzhi and Ommer, B{\"o}jrn and Navab, Nassir},
  booktitle={ICCV Workshops},
  year={2023}
}

@inproceedings{valevski2024diffusionmodelsrealtimegame,
  title={Diffusion Models Are Real-Time Game Engines},
  author={Valevski, Dani and Leviathan, Yaniv and Arar, Moab and Fruchter, Shlomi},
  booktitle={ICLR},
  year={2025}
}

@article{ha2018worldmodels,
  title={{World Models}},
  author={Ha, David and Schmidhuber, J{\"u}rgen},
  journal={arXiv preprint arXiv:1803.10122},
  year={2018}
}

@article{ha2018recurrent,
  title={Recurrent world models facilitate policy evolution},
  author={Ha, David and Schmidhuber, J{\"u}rgen},
  journal={NeurIPS},
  year={2018}
}

@misc{enigma2025multiverse,
  title={Introducing Multiverse: The First AI Multiplayer World Model},
  author={{Enigma team}},
  year={2025},
  url={https://enigma.inc/blog}
}

@inproceedings{GameNGen,
  title={Diffusion models are real-time game engines},
  author={Valevski, Dani and Leviathan, Yaniv and Arar, Moab and Fruchter, Shlomi},
  booktitle={ICLR},
  year={2025}
}

@article{GAIA-1,
  title={{GAIA-1}: A generative world model for autonomous driving},
  author={Hu, Anthony and Russell, Lloyd and Yeo, Hudson and Murez, Zak and Fedoseev, George and Kendall, Alex and Shotton, Jamie and Corrado, Gianluca},
  journal={arXiv preprint arXiv:2309.17080},
  year={2023}
}

@inproceedings{TrafficSim,
  title={{TrafficSim}: Learning to simulate realistic multi-agent behaviors},
  author={Suo, Simon and Regalado, Sebastian and Casas, Sergio and Urtasun, Raquel},
  booktitle={CVPR},
  year={2021}
}

@inproceedings{VMAS,
  title={{VMAS}: A vectorized multi-agent simulator for collective robot learning},
  author={Bettini, Matteo and Kortvelesy, Ryan and Blumenkamp, Jan and Prorok, Amanda},
  booktitle={International Symposium on Distributed Autonomous Robotic Systems},
  year={2022}
}

@article{RoboTwin2,
  title={{RoboTwin 2.0}: A scalable data generator and benchmark with strong domain randomization for robust bimanual robotic manipulation},
  author={Chen, Tianxing and Chen, Zanxin and Chen, Baijun and Cai, Zijian and Liu, Yibin and Li, Zixuan and Liang, Qiwei and Lin, Xianliang and Ge, Yiheng and Gu, Zhenyu and others},
  journal={arXiv preprint arXiv:2506.18088},
  year={2025}
}

@article{genie3,
  title         = {Genie 3: A New Frontier for World Models},
  author        = {Philip J. Ball and others},
  year          = {2025},
  url           = {}
}

@inproceedings{pondaven2025ditflow,
      title={Video Motion Transfer with Diffusion Transformers}, 
      author={Alexander Pondaven and Aliaksandr Siarohin and Sergey Tulyakov and Philip Torr and Fabio Pizzati},
      booktitle={CVPR},
      year={2025}
}

@inproceedings{VMC,
  title={{VMC}: Video motion customization using temporal attention adaption for text-to-video diffusion models},
  author={Jeong, Hyeonho and Park, Geon Yeong and Ye, Jong Chul},
  booktitle={CVPR},
  year={2024}
}

@inproceedings{MotionDirector,
  title={{MotionDirector}: Motion customization of text-to-video diffusion models},
  author={Zhao, Rui and Gu, Yuchao and Wu, Jay Zhangjie and Zhang, David Junhao and Liu, Jia-Wei and Wu, Weijia and Keppo, Jussi and Shou, Mike Zheng},
  booktitle={ECCV},
  year={2024}
}

@inproceedings{SMM,
  title={Space-time diffusion features for zero-shot text-driven motion transfer},
  author={Yatim, Danah and Fridman, Rafail and Bar-Tal, Omer and Kasten, Yoni and Dekel, Tali},
  booktitle={CVPR},
  year={2024}
}

@article{MOFT,
  title={Video diffusion models are training-free motion interpreter and controller},
  author={Xiao, Zeqi and Zhou, Yifan and Yang, Shuai and Pan, Xingang},
  journal={NeurIPS},
  year={2024}
}

@inproceedings{geng2024motionprompting,
  title={{Motion Prompting}: Controlling video generation with motion trajectories},
  author={Geng, Daniel and others},
  booktitle={CVPR},
  year={2025}
}

@inproceedings{menapace2021playablevideogeneration,
  title={Playable video generation},
  author={Menapace, Willi and Lathuiliere, Stephane and Tulyakov, Sergey and Siarohin, Aliaksandr and Ricci, Elisa},
  booktitle={CVPR},
  year={2021}
}

@inproceedings{menapace2022playableenvironmentsvideomanipulation,
  title={{Playable Environments}: Video manipulation in space and time},
  author={Menapace, Willi and Lathuiliere, St{\'e}phane and Siarohin, Aliaksandr and Theobalt, Christian and Tulyakov, Sergey and Golyanik, Vladislav and Ricci, Elisa},
  booktitle={CVPR},
  year={2022}
}

@article{agapiou2023meltingpot20,
  title={{Melting Pot} 2.0},
  author={Agapiou, John P and others},
  journal={arXiv preprint arXiv:2211.13746},
  year={2022}
}

@article{wang2026factoredlatentactionworld,
  title={Factored Latent Action World Models},
  author={Wang, Zizhao and Shi, Chang and Hu, Jiaheng and Rohling, Kevin and Mart{\'\i}n-Mart{\'\i}n, Roberto and Zhang, Amy and Stone, Peter},
  journal={arXiv preprint arXiv:2602.16229},
  year={2026}
}

@inproceedings{bruce2024geniegenerativeinteractiveenvironments,
  title={Genie: Generative interactive environments},
  author={Bruce, Jake and others},
  booktitle={ICML},
  year={2024}
}

@inproceedings{DreamerV1,
  title={{Dream to control}: Learning behaviors by latent imagination},
  author={Hafner, Danijar and Lillicrap, Timothy and Ba, Jimmy and Norouzi, Mohammad},
  booktitle={ICLR},
  year={2020}
}

@article{wang2025vdfdmultiagentvaluedecomposition,
  title={{VDFD}: Multi-Agent Value Decomposition Framework with Disentangled World Model}, 
  author={Zhizun Wang and David Meger},
  journal={arXiv preprint arXiv:2309.04615},
  year={2023}
}

@misc{xue2025learning,
title={Learning Disentangled Multi-Agent World Model for Decentralized Control},
author={Di Xue and Jing Jiang and Shaowei Zhang and Wenhao Guo and Lei Yuan and Zongzhang Zhang and Yang Yu},
year={2025},
url={https://openreview.net/forum?id=2KTEaY0lLi}
}

@article{solaris2026,
  title={Solaris: Building a Multiplayer Video World Model in Minecraft},
  author={Savva, Georgy and Michel, Oscar and Lu, Daohan and Waiwitlikhit, Suppakit and Meehan, Timothy and Mishra, Dhairya and Poddar, Srivats and Lu, Jack and Xie, Saining},
  journal={arXiv preprint arXiv:2602.22208},
  year={2026}
}

@misc{oasis2024,
  author={Decart, Etched and McIntyre, Quinn and Campbell, Spruce and Chen, Xinlei and Wachen, Robert},
  title={Oasis: A Universe in a Transformer},
  year={2024},
  url={https://oasis-model.github.io/}
}

@article{WorldMem,
  title={{WorldMem}: Long-term consistent world simulation with memory},
  author={Xiao, Zeqi and Lan, Yushi and Zhou, Yifan and Ouyang, Wenqi and Yang, Shuai and Zeng, Yanhong and Pan, Xingang},
  journal={NeurIPS},
  year={2025}
}

@article{DDPMWorldModel,
  title={Diffusion for world modeling: Visual details matter in atari},
  author={Alonso, Eloi and Jelley, Adam and Micheli, Vincent and Kanervisto, Anssi and Storkey, Amos J and Pearce, Tim and Fleuret, Fran{\c{c}}ois},
  journal={NeurIPS},
  year={2024}
}

@article{zhang2025matrixgameinteractiveworldfoundation,
  title={{Matrix-Game}: Interactive world foundation model},
  author={Zhang, Yifan and others},
  journal={arXiv preprint arXiv:2506.18701},
  year={2025}
}

@article{he2025matrixgame20opensourcerealtime,
  title={{Matrix-Game 2.0}: An open-source real-time and streaming interactive world model},
  author={He, Xianglong and others},
  journal={arXiv preprint arXiv:2508.13009},
  year={2025}
}

@inproceedings{yu2025gamefactorycreatingnewgames,
  title={{GameFactory}: Creating new games with generative interactive videos},
  author={Yu, Jiwen and Qin, Yiran and Wang, Xintao and Wan, Pengfei and Zhang, Di and Liu, Xihui},
  booktitle={ICCV},
  year={2025}
}

@article{RELIC,
  title={{RELIC}: Interactive video world model with long-horizon memory},
  author={Hong, Yicong and others},
  journal={arXiv preprint arXiv:2512.04040},
  year={2025}
}

@article{wang2024boximatorgeneratingrichcontrollable,
  title={Boximator: Generating rich and controllable motions for video synthesis},
  author={Wang, Jiawei and Zhang, Yuchen and Zou, Jiaxin and Zeng, Yan and Wei, Guoqiang and Yuan, Liping and Li, Hang},
  journal={arXiv preprint arXiv:2402.01566},
  year={2024}
}

@article{FVD,
  title={Towards accurate generative models of video: A new metric \& challenges},
  author={Unterthiner, Thomas and van Steenkiste, Sjoerd and Kurach, Karol and Marinier, Raphael and Michalski, Marcin and Gelly, Sylvain},
  journal={arXiv preprint arXiv:1812.01717},
  year={2018}
}

@inproceedings{LPIPS,
  title={The Unreasonable Effectiveness of Deep Features as a Perceptual Metric},
  author={Zhang, Richard and Isola, Phillip and Efros, Alexei A and Shechtman, Eli and Wang, Oliver},
  booktitle={CVPR},
  year={2018}
}

@inproceedings{DDPM,
  title={Denoising diffusion probabilistic models},
  author={Ho, Jonathan and Jain, Ajay and Abbeel, Pieter},
  booktitle=NeurIPS,
  year={2020}
}

@article{Adam,
  title={Adam: A method for stochastic optimization},
  author={Kingma, Diederik P and Ba, Jimmy},
  journal={arXiv preprint arXiv:1412.6980},
  year={2014}
}

@inproceedings{DiffusionModels,
  title={Deep unsupervised learning using nonequilibrium thermodynamics},
  author={Sohl-Dickstein, Jascha and Weiss, Eric and Maheswaranathan, Niru and Ganguli, Surya},
  booktitle={ICML},
  year={2015}
}

@inproceedings{FlowMatching,
  title={Flow matching for generative modeling},
  author={Lipman, Yaron and Chen, Ricky TQ and Ben-Hamu, Heli and Nickel, Maximilian and Le, Matt},
  booktitle={ICLR},
  year={2023}
}

@inproceedings{RFSampler,
  title={Flow Straight and Fast: Learning to Generate and Transfer Data with Rectified Flow},
  author={Liu, Xingchao and others},
  booktitle={ICLR},
  year={2023}
}

@inproceedings{AnimateDiff,
  title={Animatediff: Animate your personalized text-to-image diffusion models without specific tuning},
  author={Guo, Yuwei and Yang, Ceyuan and Rao, Anyi and Wang, Yaohui and Qiao, Yu and Lin, Dahua and Dai, Bo},
  booktitle={ICLR},
  year={2024}
}

@article{ImagenVideo,
  title={{Imagen Video}: High definition video generation with diffusion models},
  author={Ho, Jonathan and others},
  journal={arXiv preprint arXiv:2210.02303},
  year={2022}
}

@article{MetaMovieGen,
  title={{Movie Gen}: A Cast of Media Foundation Models},
  author={Polyak, Adam and others},
  journal={arXiv preprint arXiv:2410.13720},
  year={2024}
}

@article{HunyuanVideo,
  title={{HunyuanVideo}: A systematic framework for large video generative models},
  author={Kong, Weijie and others},
  journal={arXiv preprint arXiv:2412.03603},
  year={2024}
}

@article{Wan21,
  title={{Wan}: Open and advanced large-scale video generative models},
  author={Wang, Ang and others},
  journal={arXiv preprint arXiv:2503.20314},
  year={2025}
}

@article{Step-Video-T2V,
  title={{Step-Video-T2V} technical report: The practice, challenges, and future of video foundation model},
  author={Ma, Guoqing and others},
  journal={arXiv preprint arXiv:2502.10248},
  year={2025}
}

@article{VideoDiffusionModels,
    title={Video diffusion models},
    author={Ho, Jonathan and Salimans, Tim and Gritsenko, Alexey and Chan, William and Norouzi, Mohammad and Fleet, David J},
    journal={NeurIPS},
    year={2022}
}

@inproceedings{AlignYourLatents,
  title={{Align your Latents}: High-resolution video synthesis with latent diffusion models},
  author={Blattmann, Andreas and Rombach, Robin and Ling, Huan and Dockhorn, Tim and Kim, Seung Wook and Fidler, Sanja and Kreis, Karsten},
  booktitle={CVPR},
  year={2023}
}

@inproceedings{U-Net,
  title={{U-Net}: Convolutional networks for biomedical image segmentation},
  author={Ronneberger, Olaf and Fischer, Philipp and Brox, Thomas},
  booktitle={MICCAI},
  year={2015}
}

@article{SnapVideo,
  title={Snap Video: Scaled Spatiotemporal Transformers for Text-to-Video Synthesis},
  author={Menapace, Willi and others},
  journal={CVPR},
  year={2024}
}

@inproceedings{GLIGEN,
  title={{GLIGEN}: Open-set grounded text-to-image generation},
  author={Li, Yuheng and Liu, Haotian and Wu, Qingyang and Mu, Fangzhou and Yang, Jianwei and Gao, Jianfeng and Li, Chunyuan and Lee, Yong Jae},
  booktitle={CVPR},
  year={2023}
}

@inproceedings{LLMGroundedDM,
  title={Llm-grounded video diffusion models},
  author={Lian, Long and Shi, Baifeng and Yala, Adam and Darrell, Trevor and Li, Boyi},
  booktitle={ICLR},
  year={2024}
}

@article{VideoCrafter1,
  title={{VideoCrafter1}: Open Diffusion Models for High-Quality Video Generation}, 
  author={Haoxin Chen and Menghan Xia and Yingqing He and Yong Zhang and Xiaodong Cun and Shaoshu Yang and Jinbo Xing and Yaofang Liu and Qifeng Chen and Xintao Wang and Chao Weng and Ying Shan},
  year={2023},
  journal={arXiv preprint arXiv:2310.19512},
}

@inproceedings{VideoCrafter2,
  title={{VideoCrafter2}: Overcoming Data Limitations for High-Quality Video Diffusion Models}, 
  author={Haoxin Chen and Yong Zhang and Xiaodong Cun and Menghan Xia and Xintao Wang and Chao Weng and Ying Shan},
  booktitle={CVPR},
  year={2024}
}

@inproceedings{DynamiCrafter,
  title={{DynamiCrafter}: Animating open-domain images with video diffusion priors},
  author={Xing, Jinbo and Xia, Menghan and Zhang, Yong and Chen, Haoxin and Wang, Xintao and Wong, Tien-Tsin and Shan, Ying},
  booktitle={ECCV},
  year={2024}
}

@misc{Seaweed-7B,
  title={{Seaweed-7B}: Cost-Effective Training of Video Generation Foundation Model}, 
  author={Team Seawead and Ceyuan Yang and others},
  journal={arXiv preprint arXiv:2504.08685},
  year={2025}
}

@article{Sora,
  title={Video generation models as world simulators},
  author={Tim Brooks and others},
  year={2024},
  url={https://openai.com/research/video-generation-models-as-world-simulators},
  journal={OpenAI technical reports}
}

@misc{GoogleVeo,
  title={Veo},
  author = {Abhishek Sharma and others},
  year={2024},
  url={https://deepmind.google/technologies/veo/},
}

@article{SVD, 
  title={Stable video diffusion: Scaling latent video diffusion models to large datasets},
  author={Blattmann, Andreas and others},
  journal={arXiv preprint arXiv:2311.15127},
  year={2023}
}

@article{peebles2023scalablediffusionmodelstransformers,
  title={Scalable diffusion models with transformers},
  author={Peebles, William and Xie, Saining},
  journal={ICCV},
  year={2023}
}

@inproceedings{CogVideoX,
  title={{CogVideoX}: Text-to-video diffusion models with an expert transformer},
  author={Yang, Zhuoyi and others},
  booktitle={ICLR},
  year={2025}
}

@article{WALT,
  title={Photorealistic video generation with diffusion models},
  author={Gupta, Agrim and Yu, Lijun and Sohn, Kihyuk and Gu, Xiuye and Hahn, Meera and Fei-Fei, Li and Essa, Irfan and Jiang, Lu and Lezama, Jos{\'e}},
  journal={arXiv preprint arXiv:2312.06662},
  year={2023}
}

@inproceedings{SD3,
  title={Scaling rectified flow transformers for high-resolution image synthesis},
  author={Esser, Patrick and others},
  booktitle={ICML},
  year={2024}
}

@article{DragNUWA,
  title={{DragNUWA}: Fine-grained control in video generation by integrating text, image, and trajectory},
  author={Yin, Shengming and Wu, Chenfei and Liang, Jian and Shi, Jie and Li, Houqiang and Ming, Gong and Duan, Nan},
  journal={arXiv preprint arXiv:2308.08089},
  year={2023}
}

@inproceedings{DragAnything,
  title={{DragAnything}: Motion Control for Anything using Entity Representation},
  author={Wu, Wejia and Li, Zhuang and Gu, Yuchao and Zhao, Rui and He, Yefei and Zhang, David Junhao and Shou, Mike Zheng and Li, Yan and Gao, Tingting and Zhang, Di},
  booktitle={ECCV},
  year={2024}
}

@inproceedings{Tora,
  title={Tora: Trajectory-oriented diffusion transformer for video generation},
  author={Zhang, Zhenghao and Liao, Junchao and Li, Menghao and Dai, Zuozhuo and Qiu, Bingxue and Zhu, Siyu and Qin, Long and Wang, Weizhi},
  booktitle={CVPR},
  year={2025}
}

@inproceedings{SG-I2V,
  title={{SG-I2V}},
  author={Namekata, Koichi and Bahmani, Sherwin and Wu, Ziyi and Kant, Yash and Gilitschenski, Igor and Lindell, David B},
  booktitle={ICLR},
  year={2025}
}

@article{chen2024diffusionforcingnexttokenprediction,
  title={{Diffusion Forcing}: Next-token prediction meets full-sequence diffusion},
  author={Chen, Boyuan and Mart{\'\i} Mons{\'o}, Diego and Du, Yilun and Simchowitz, Max and Tedrake, Russ and Sitzmann, Vincent},
  journal={NeurIPS},
  year={2024}
}

@inproceedings{DMD,
  title={One-step diffusion with distribution matching distillation},
  author={Yin, Tianwei and Gharbi, Micha{\"e}l and Zhang, Richard and Shechtman, Eli and Durand, Fredo and Freeman, William T and Park, Taesung},
  booktitle={CVPR},
  year={2024}
}

@article{DMDv2,
  title={Improved distribution matching distillation for fast image synthesis},
  author={Yin, Tianwei and Gharbi, Micha{\"e}l and Park, Taesung and Zhang, Richard and Shechtman, Eli and Durand, Fredo and Freeman, Bill},
  journal={NeurIPS},
  year={2024}
}

@inproceedings{CausVid,
  title={From slow bidirectional to fast autoregressive video diffusion models},
  author={Yin, Tianwei and Zhang, Qiang and Zhang, Richard and Freeman, William T and Durand, Fredo and Shechtman, Eli and Huang, Xun},
  booktitle={CVPR},
  year={2025}
}

@article{SelfForcing,
  title={{Self Forcing}: Bridging the Train-Test Gap in Autoregressive Video Diffusion},
  author={Huang, Xun and Li, Zhengqi and He, Guande and Zhou, Mingyuan and Shechtman, Eli},
  journal={NeurIPS},
  year={2025}
}

@inproceedings{MinT,
  title={{Mind the Time}: Temporally-Controlled Multi-Event Video Generation},
  author={Wu, Ziyi and Siarohin, Aliaksandr and Menapace, Willi and Skorokhodov, Ivan and Fang, Yuwei and Chordia, Varnith and Gilitschenski, Igor and Tulyakov, Sergey},
  booktitle={CVPR},
  year={2025}
}

@inproceedings{VideoDirectorGPT,
  title={{VideoDirectorGPT}: Consistent multi-scene video generation via llm-guided planning},
  author={Lin, Han and Zala, Abhay and Cho, Jaemin and Bansal, Mohit},
  booktitle={COLM},
  year={2024}
}

@article{su2023roformerenhancedtransformerrotary,
  title={Roformer: Enhanced transformer with rotary position embedding},
  author={Su, Jianlin and Ahmed, Murtadha and Lu, Yu and Pan, Shengfeng and Bo, Wen and Liu, Yunfeng},
  journal={Neurocomputing},
  year={2024}
}

@article{FlexAttention,
  title={{Flex Attention}: A programming model for generating optimized attention kernels},
  author={Dong, Juechu and Feng, Boyuan and Guessous, Driss and Liang, Yanbo and He, Horace},
  journal={arXiv preprint arXiv:2412.05496},
  year={2024}
}

@inproceedings{MatchDiffusion,
  title = {MatchDiffusion: Training-free Generation of Match-Cuts},
  author = {Pardo, Alejandro and Pizzati, Fabio and Zhang, Tong and Pondaven, Alexander and Torr, Philip and Perez, Juan Camilo and Ghanem, Bernard},
  booktitle = {ICCV},
  year = {2025},
}

@inproceedings{tevet2022humanmotiondiffusionmodel,
  title={Human motion diffusion model},
  author={Tevet, Guy and Raab, Sigal and Gordon, Brian and Shafir, Yonatan and Cohen-Or, Daniel and Bermano, Amit H},
  booktitle={ICLR},
  year={2023}
}

@article{zhang2022motiondiffusetextdrivenhumanmotion,
  title={Motiondiffuse: Text-driven human motion generation with diffusion model},
  author={Zhang, Mingyuan and Cai, Zhongang and Pan, Liang and Hong, Fangzhou and Guo, Xinying and Yang, Lei and Liu, Ziwei},
  journal={PAMI},
  year={2024}
}

@article{jiang2023motiongpthumanmotionforeign,
  title={{MotionGPT}: Human motion as a foreign language},
  author={Jiang, Biao and Chen, Xin and Liu, Wen and Yu, Jingyi and Yu, Gang and Chen, Tao},
  journal={NeurIPS},
  year={2023}
}

@inproceedings{chi2024m2d2mmultimotiongenerationtext,
  title={{M2D2M}: Multi-motion generation from text with discrete diffusion models},
  author={Chi, Seunggeun and Chi, Hyung-gun and Ma, Hengbo and Agarwal, Nakul and Siddiqui, Faizan and Ramani, Karthik and Lee, Kwonjoon},
  booktitle={ECCV},
  year={2024}
}

@inproceedings{guidedmotion,
  title={Local Action-Guided Motion Diffusion Model for Text-to-Motion Generation},
  author={Jin, Peng and Li, Hao and Cheng, Zesen and Li, Kehan and Yu, Runyi and Liu, Chang and Ji, Xiangyang and Yuan, Li and Chen, jie},
  booktitle={ECCV},
  year={2024}
}

@article{gokmen2025ropecrafttrainingfreemotiontransfer,
  title={{RoPECraft}: Training-Free Motion Transfer with Trajectory-Guided RoPE Optimization on Diffusion Transformers},
  author={Gokmen, Ahmet Berke and Ekin, Yigit and Bilecen, Bahri Batuhan and Dundar, Aysegul},
  journal={NeurIPS},
  year={2025}
}

@article{rassin2024linguisticbindingdiffusionmodels,
  title={{Linguistic Binding in Diffusion Models}: Enhancing attribute correspondence through attention map alignment},
  author={Rassin, Royi and Hirsch, Eran and Glickman, Daniel and Ravfogel, Shauli and Goldberg, Yoav and Chechik, Gal},
  journal={NeurIPS},
  year={2023}
}

@article{greff2020bindingproblemartificialneural,
  title={On the binding problem in artificial neural networks},
  author={Greff, Klaus and Van Steenkiste, Sjoerd and Schmidhuber, J{\"u}rgen},
  journal={arXiv preprint arXiv:2012.05208},
  year={2020}
}

@article{zarei2025improvingcompositionalattributebinding,
  title={Improving compositional attribute binding in text-to-image generative models via enhanced text embeddings},
  author={Zarei, Arman and Rezaei, Keivan and Basu, Samyadeep and Saberi, Mehrdad and Moayeri, Mazda and Kattakinda, Priyatham and Feizi, Soheil},
  journal={arXiv preprint arXiv:2406.07844},
  year={2024}
}

@article{chefer2023attendandexciteattentionbasedsemanticguidance,
  title={{Attend-and-Excite}: Attention-based semantic guidance for text-to-image diffusion models},
  author={Chefer, Hila and Alaluf, Yuval and Vinker, Yael and Wolf, Lior and Cohen-Or, Daniel},
  journal={TOG},
  year={2023}
}

@inproceedings{liu2023compositionalvisualgenerationcomposable,
  title={Compositional visual generation with composable diffusion models},
  author={Liu, Nan and Li, Shuang and Du, Yilun and Torralba, Antonio and Tenenbaum, Joshua B},
  booktitle={ECCV},
  year={2022}
}

@article{t2icompbenchpp,
  title={{T2I-CompBench++}: An enhanced and comprehensive benchmark for compositional text-to-image generation},
  author={Huang, Kaiyi and Duan, Chengqi and Sun, Kaiyue and Xie, Enze and Li, Zhenguo and Liu, Xihui},
  journal={PAMI},
  year={2025}
}

@article{NeuralAssets,
  title={{Neural Assets}: 3d-aware multi-object scene synthesis with image diffusion models},
  author={Wu, Ziyi and Rubanova, Yulia and Kabra, Rishabh and Hudson, Drew A and Gilitschenski, Igor and Aytar, Yusuf and Van Steenkiste, Sjoerd and Allen, Kelsey R and Kipf, Thomas},
  journal={NeurIPS},
  year={2024}
}

@article{multigen,
      title={MultiGen: Level-Design for Editable Multiplayer Worlds in Diffusion Game Engines},
      author={Ryan Po and David Junhao Zhang and Amir Hertz and Gordon Wetzstein and Neal Wadhwa and Nataniel Ruiz},
      journal={arXiv preprint arXiv:2603.06679},
      year={2026},
}
